# Medi-Gemma: A Hybrid Clinical Decision Support System Integrating Deterministic EMR Analytics and Retrieval-Augmented Generation

Mohammed Saim Ahmed Quadri*, Yunzhe Xue†, Justin W. Ady‡, and Usman Roshan†
* Department of Computer Science, New Jersey Institute of Technology
Newark, NJ, USA
mohammedsaimquadri@gmail.com
† Department of Data Science, New Jersey Institute of Technology
Newark, NJ, USA
yunzhexue@gmail.com; usmanroshan@gmail.com
‡ Vascular and Endovascular Surgery, Robert Wood Johnson Hospital
New Brunswick, NJ, USA
jwa60@rwjms.rutgers.edu

***Abstract*—Deploying Large Language Models (LLMs) in high-stakes clinical settings remains limited by structural hallucinations, weak deterministic reasoning over tabular patient data, and omissions in vector retrieval. This paper presents the architecture and validation of Medi-Gemma, a Clinical Decision Support System (CDSS) for wound pathology triage and workflow automation. The platform introduces a decoupled framework that separates clinical perception from data orchestration while preserving traceable reasoning. Medi-Gemma uses a multi-stage pipeline coordinated by a centralized ClinicalOrchestrator. Data requests are handled without generative inference by a DataManager that cleans unstructured Electronic Medical Record (EMR) files through type coercion. Natural language queries are processed by a hierarchical IntentRouter, which routes requests to deterministic analytics paths executed by a PandasQueryEngine or to patient-specific reasoning managed by a ClinicalRAGEngine using a CPU-optimized vector store. A key contribution is the Ground Truth Injection Module, which intercepts patient-specific queries, extracts numeric identification tokens, queries the structured dataframe via Pandas, retrieves the latest validated clinical state, and embeds this snapshot as an overriding context block in the LLM prompt before generation. Safety compliance is enforced by a deterministic ProtocolManager that maps clinical terminology to fixed evidence-based risk pathways, while a SafetyVerifier phrase filter prevents output rule violations. Validation shows that this architecture eliminates semantic context drift, prevents database compilation crashes, and improves factual adherence to backend clinical repositories. These results support Medi-Gemma as a safer pattern for LLM-based clinical decision support where structured data fidelity, retrieval grounding, and deterministic safeguards are essential.**

## I. Introduction

The integration of Large Language Models (LLMs) into hospital workflows holds transformative potential for reducing clinician documentation overhead, generating structured Subjective, Objective, Assessment, and Plan (SOAP) progress notes, and accelerating patient risk stratification [1], [2], [4], [5]. However, standard generative text architectures remain highly vulnerable to errors that limit their safe application in clinical environments. When processing unstructured narrative chart notes alongside structured longitudinal datasets (such as vital sign trends, lab results, and wound size dimensions), standard architectures frequently fail.

These vulnerabilities stem from two primary architectural limitations that have also been identified in recent work on retrieval augmented medical language models and clinical AI systems [3], [5], [8]:

- **Vector Retrieval Omissions:** Standard Retrieval-Augmented Generation (RAG) models split text files into arbitrary chunks and retrieve data based purely on semantic similarity scores. In longitudinal patient tracking, this can cause the system to overlook critical historical context or prioritize an outdated chart note over the patient's current clinical state.
- **Tabular Hallucinations:** Generative models are inherently poor at performing exact quantitative operations over structured data tables. For example, asking a standard LLM to "calculate the average wound healing rate for a diabetic cohort" often results in fabricated metrics, incorrect patient filtering, or database processing errors.

To address these challenges, we present Medi-Gemma, a hybrid Clinical Decision Support System (CDSS) that combines deterministic computation with retrieval-augmented generation for trustworthy clinical reasoning, building upon recent developments in hybrid medical AI systems [7], [8]. The core innovation centers on a centralized `ClinicalOrchestrator` that utilizes dynamic intent routing alongside deterministic ground truth context injection. By programmatically enforcing data types on incoming records, isolating cohort statistics within a code-execution engine, and embedding the patient's absolute latest clinical measurements directly into the prompt envelope, the system eliminates semantic context drift and guarantees factual adherence to the patient record.

The primary contributions of this work are summarized as follows:

- A hybrid Clinical Decision Support System that explicitly separates deterministic computation from generative reasoning.
- A dynamic intent routing framework that directs analytical and patient-specific queries to specialized execution pipelines.
- A Ground Truth State Injection mechanism that deterministically incorporates the latest validated patient information prior to language model inference.
- A modular clinical architecture integrating retrieval-augmented generation, protocol mapping, and rule-based safety verification for trustworthy clinical decision support.

## II. Related Work

Recent advances in Large Language Models (LLMs) have accelerated the development of intelligent Clinical Decision Support Systems (CDSSs) capable of assisting clinicians with documentation, summarization, patient monitoring, and medical reasoning [1], [2], [4], [5]. Retrieval-Augmented Generation (RAG) has emerged as one of the dominant paradigms for reducing hallucinations by grounding language models in external knowledge sources rather than relying solely on parametric memory [3], [16].

Several medical LLMs, including Med-PaLM, LLaVA-Med, and other clinically adapted transformer architectures, have demonstrated encouraging performance on medical question answering and multimodal reasoning tasks [4]–[6], [9], [17]. However, most existing systems primarily retrieve unstructured textual information and rely almost entirely on the language model to integrate retrieved context into final responses.

Clinical Electronic Medical Records (EMRs), however, contain a substantial amount of structured longitudinal information including wound measurements, laboratory values, encounter histories, medications, and provider assessments. Recent foundation-model based clinical agents have demonstrated improved medical reasoning through modular planning and tool use; however, few systems explicitly integrate deterministic database operations into the reasoning pipeline, leaving structured numerical information vulnerable to retrieval inconsistencies. Conventional RAG pipelines frequently treat these structured records as plain text, making quantitative reasoning, temporal comparisons, and cohort-level analytics susceptible to hallucination and retrieval inconsistencies [3], [7].

Recent work has therefore explored hybrid clinical AI systems, agentic reasoning frameworks, and tool-augmented language models that combine deterministic computation with foundation models. Representative examples include ReAct, Toolformer, and HuggingGPT, which demonstrate that external tool invocation substantially improves reasoning reliability and factual consistency by allowing language models to interact with specialized computational modules rather than relying solely on parametric knowledge [8], [18]–[20].

Our work extends this direction by introducing a modular clinical agent architecture that explicitly separates deterministic EMR analytics from semantic retrieval and generative reasoning while incorporating evidence-based safety verification. Unlike general-purpose agent frameworks, the proposed system is specifically designed for structured clinical workflows in which deterministic patient-state retrieval precedes language model inference [15], [18], [20], [21]. Rather than asking a language model to infer the current patient state from retrieved documents alone, the architecture injects the most recent validated clinical measurements directly into the prompt, substantially reducing temporal inconsistencies while preserving explainability and auditability.

TABLE I
Comparison between conventional clinical RAG systems and the proposed Medi-Gemma architecture.

| **Capability** | Conventional RAG | **Medi-Gemma** |
|---|---|---|
| Semantic Retrieval | ✓ | ✓ |
| Structured EMR Analytics | – | ✓ |
| Dynamic Intent Routing | – | ✓ |
| Ground Truth State Injection | – | ✓ |
| Protocol-based Reasoning | Limited | ✓ |
| Safety Verification Layer | Limited | ✓ |
| Dashboard Analytics | – | ✓ |
| Multimodal Image Analysis | Limited | ✓ |

## III. System Architecture

### *A. EMR Data Preprocessing and Ingestion*

The system ingestion pipeline is driven by a specialized `DataManager` designed to process unstructured or non-standard hospital encounter files and patient demographics sheets. This preprocessing strategy follows the deterministic data processing philosophy commonly adopted in modern scientific computing frameworks such as Pandas [15]. A major source of system instability in clinical data processing is the presence of missing elements, unformatted string markers, or arbitrary text values (such as "N/A" or "unknown") within quantitative fields. To prevent backend database compilation crashes and ensure data integrity, the ingestion engine passes uploaded file streams through an aggressive type-coercion pipeline.

The data pipeline standardizes varying file formats by mapping common synonyms (such as `patient_id`) to a single uniform key. The ingestion array then isolates key metrics including `Wound_Size_Length_cm`, `Wound_Size_Width_cm`, `Wound_Size_Area_cm2`, `Necrosis_Percent`, `Slough_Percent`, `Granulation_Percent`, and `Pain_Level`—and converts them to uniform floats using Pandas numeric coercion. Non-numeric string errors are safely converted to distinct null entries.

To support descriptive clinical dashboard generation without data loss, these quantitative null values are filled with a default of 0.0. Conversely, text columns (such as `Narrative`, `Notes`, `Treatment_Plan`, and `Exudate_Type`) retain missing entries as distinct string markers, ensuring the system can properly flag unrecorded patient information during downstream data gap analysis.

### *B. Dynamic Intent Routing*

Once the data is cleaned and structured, the runtime query loop is governed by a decoupled classification router. When a clinician inputs a natural language prompt, the `IntentRouter` parses the linguistic tokens before invoking any heavy text-generation pipelines [15]. This classification process follows a strict rule-based priority hierarchy:

- **Rule 1 (Specific Patient Target):** The system scans the prompt for specific patient identifiers using regular expressions matching terms like `patient`, `id`, or `encounter` followed by a numeric sequence. If found, the query is automatically routed to the **Clinical Route** (`QueryIntent.CLINICAL`), bypassing global table lookups to prevent cross-patient context leaks.
- **Rule 2 (Statistical Aggregation):** If no individual identifier is found, the engine scans the prompt for data aggregation keywords, including *how many*, *count*, *total*, *average*, *mean*, *percentage*, *breakdown*, or *census*. If any of these tokens match, the query is routed to the **Data Route** (`QueryIntent.DATA`).
- **Rule 3 (Visual / Follow-up Tracking):** If the query relies on contextual references (such as *this wound* or *the eschar patch*) following an active image upload, it is routed to the **Clinical Route** to maintain continuity of care over the current session.

### *C. Ground Truth State Injection and Clinical Orchestration*

When a query is classified under the data analytics track, it is routed directly to the `PandasQueryEngine`. Backed by a local instance of Gemma via Ollama, this engine interprets the query text and generates executable Python code to manipulate the underlying dataframe. This process uses strict prompt constraints that force the system to perform exact programmatic operations (such as `.mean()` or `.count()`) directly against the data frame rows. This ensures that quantitative questions are answered with mathematical precision, completely avoiding the risks of arithmetic hallucinations common in standard text models [15].

For patient-specific clinical lookups, the system invokes the `ClinicalOrchestrator` to execute a multi-tier **Ground Truth State Injection** process:

1) **Identifier Extraction:** The orchestrator extracts numeric target tokens from the query text and validates them against the active patient registry to ensure the record exists.
2) **Pandas State Extraction:** The engine bypasses vector index querying to perform a direct look-up on the structured dataframe, filtering by the validated patient ID. It sorts the historical encounters by `Encounter_Date` to isolate the patient's absolute latest clinical encounter.
3) **Prompt Envelope Enrichment:** The orchestrator extracts key clinical parameters—including the precise date, physical wound dimensions, localized tissue composition percentages, and the last recorded nurse notes and constructs a verified context block.
4) **Enforcement Instruction:** This context block is injected into the model's prompt window with explicit enforcement commands: `[SYSTEM INJECTED GROUND TRUTH - PRIORITY OVER RETRIEVAL] ... You MUST use this date as the current status.`

### *D. Semantic Vector Retrieval Engine*

Descriptive narrative tracking and long-term trend summaries are processed by the `ClinicalRAGEngine`. Longitudinal text encounters are split into structured blocks and converted to embeddings using a locally deployed `BAAI/bge-small-en-v1.5` model running entirely on CPU to ensure data privacy. The retrieval pipeline is implemented using LlamaIndex over a vector similarity index [11], [12]. These text blocks are stored within a LlamaIndex vector store using structured metadata attachments that bind each chunk to its respective identifier.

When executing semantic searches, the engine applies a strict metadata overlay using `MetadataFilters` and an `ExactMatchFilter` configured to match the target patient ID. This restricts the vector database search space, ensuring the retriever only extracts text chunks belonging to that specific patient. This metadata restriction prevents cross-contamination of record context across different patients. The retrieved chunks are then combined with the injected ground truth state to provide a rich context window for localized inference.

### *E. Deterministic Clinical Safety Layer*

To enforce clinical safety limits and eliminate linguistic ambiguity, the finalized text context is evaluated through a deterministic, rule-based safety framework. Raw text generated during model inference is checked by the `ProtocolManager`, which maps identified pathology terms to a structured severity hierarchy defined in `config/protocols.yaml`:

This keyword matching hierarchy prioritizes life-threatening conditions (such as necrotizing fasciitis or active infection markers), ensuring the system defaults to the highest applicable risk tier regardless of any mitigating text formatting. Additionally, outputs are passed through the `SafetyVerifier` keyword filter. This filter scans for banned or hazardous phrases (e.g., specific unauthorized dosage instructions or attempts to ignore system guidelines), replacing unsafe generations with an automated safety alert before they can be displayed to the clinician.

## IV. System Validation

The proposed Clinical Decision Support System (CDSS) was validated through representative clinical workflows designed to evaluate the interaction between deterministic EMR analytics, retrieval-augmented generation, dynamic intent routing, and rule-based safety enforcement. Rather than benchmarking standalone language models, the objective of this validation was to verify that each architectural component performed its intended function while maintaining factual consistency with structured Electronic Medical Records (EMRs) [7], [8].

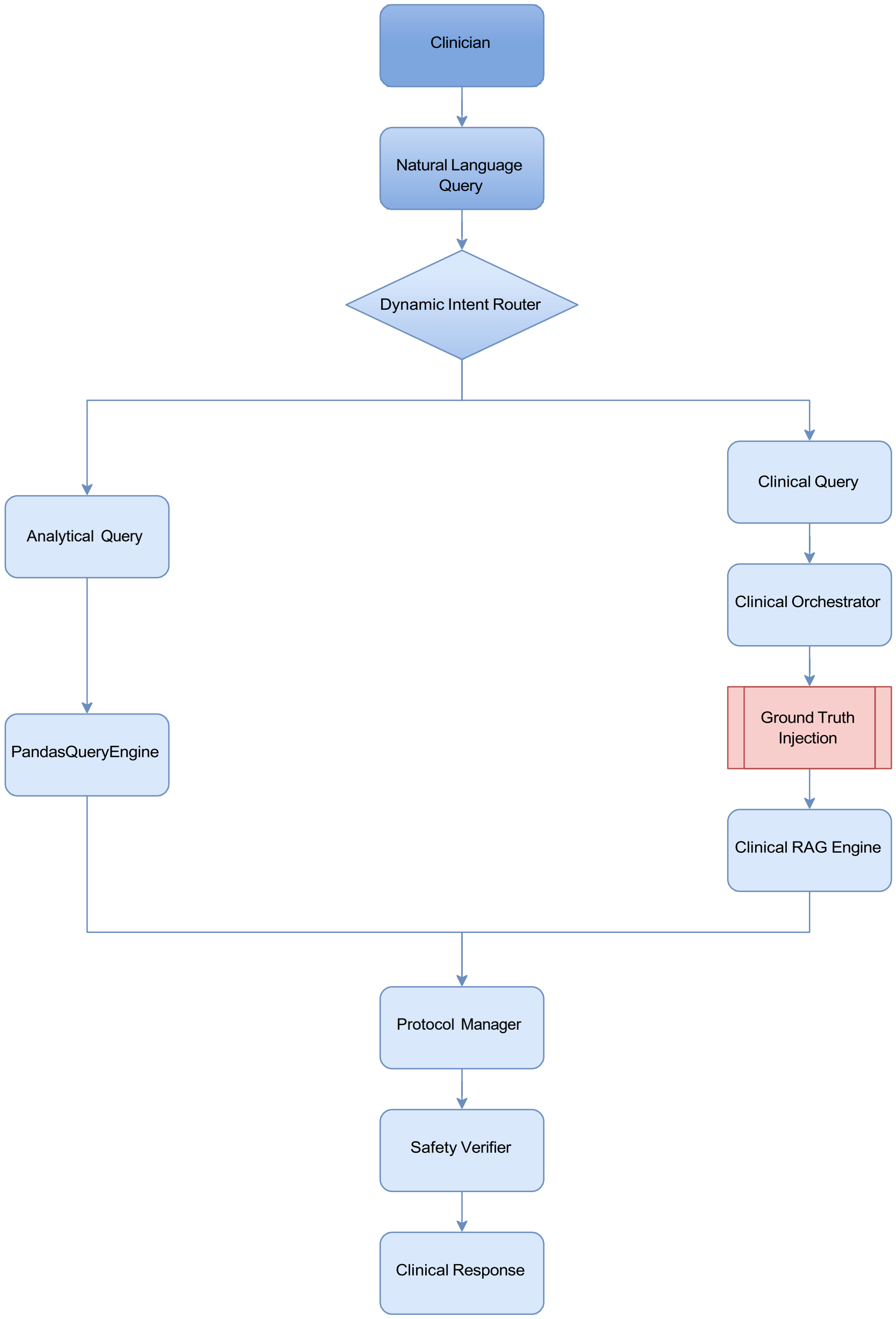

Fig. 1. Overall architecture of the proposed Medi-Gemma Clinical Decision Support System. Incoming clinician queries are dynamically routed to deterministic analytics or patient-specific reasoning pipelines. The Clinical Orchestrator performs Ground Truth State Injection before retrieval-augmented generation, while the Protocol Manager and Safety Verifier enforce evidence-based clinical safety prior to response generation.

The validation focused on four key system capabilities:

- Reliable ingestion and normalization of heterogeneous EMR datasets.

TABLE II
EVIDENCE-BASED CLINICAL PROTOCOL MAPPING USED BY THE RULE-BASED SAFETY LAYER

| Protocol ID Key | Risk Level | Target Code Trigger Synonyms |
|---|---|---|
| critical_infection | Critical Risk | dishwater, crepitus, necrotizing |
| infection_suspected | High Risk | pus, purulent, abscess, cellulitis, odor |
| pressure_injury | High Risk | eschar, black tissue, necrotic, slough |
| diabetic_foot_ulcer | High Risk | diabetic, DFU, plantar, neuropathic |
| surgical_wound | Medium Risk | surgical, incision, suture, dehiscence |

- Accurate routing of clinician queries between deterministic analytics and retrieval-based reasoning.
- Deterministic Ground Truth State Injection for patient-specific reasoning.
- Rule-based protocol enforcement and safety verification prior to response generation.

TABLE III
CORE COMPONENTS OF THE PROPOSED CLINICAL DECISION SUPPORT SYSTEM

| Component | Primary Function |
|---|---|
| DataManager | EMR ingestion and preprocessing |
| IntentRouter | Query classification |
| PandasQueryEngine | Deterministic analytics |
| ClinicalOrchestrator | Workflow coordination |
| Ground Truth Injection | Latest patient-state retrieval |
| ClinicalRAGEngine | Context retrieval |
| ProtocolManager | Guideline mapping |
| SafetyVerifier | Output safety validation |

### A. EMR Data Processing Validation

The DataManager was evaluated using representative encounter and patient datasets containing both structured numerical measurements and unstructured clinical documentation. Incoming files contained heterogeneous column naming conventions, missing numerical entries, inconsistent formatting, and mixed data types representative of real-world EMR exports. During ingestion, synonymous field names were standardized into a unified internal schema while numerical wound measurements were converted through strict type coercion. Missing numerical values were safely represented using null-aware defaults without interrupting downstream processing. Textual narrative fields preserved missing values as explicit placeholders, enabling the system to distinguish incomplete documentation from absent pathology.

These preprocessing steps ensured that deterministic analytics, semantic retrieval, and language generation operated on a consistent and validated patient representation.

### B. Dynamic Intent Routing Validation

The IntentRouter was validated using representative clinician queries spanning statistical analysis, patient-specific reasoning, and follow-up wound assessment. Each incoming query was classified before any language model invocation, ensuring that deterministic database operations were isolated from generative reasoning.

Queries requesting cohort-level statistics, averages, patient counts, or percentage calculations were routed directly to the PandasQueryEngine, where executable Python operations were performed on structured EMR tables. In contrast, patient-specific questions involving wound progression, treatment planning, or encounter summaries were directed to the ClinicalRAGEngine.

This routing strategy prevented unnecessary retrieval and language generation for purely analytical questions while ensuring that patient-specific reasoning always incorporated contextual clinical information.

### C. Ground Truth State Injection Validation

A primary contribution of the proposed architecture is the Ground Truth State Injection mechanism. Conventional retrieval-augmented generation systems rely exclusively on semantic similarity retrieval, making them susceptible to outdated encounter summaries or incomplete contextual retrieval. This limitation has been widely discussed in the RAG literature, where retrieval quality directly influences factual consistency [3].

In the proposed framework, patient identifiers are first extracted directly from clinician queries using deterministic parsing. The ClinicalOrchestrator then performs an exact lookup within the structured EMR dataframe to retrieve the patient's most recent documented encounter. Clinical measurements including wound dimensions, tissue composition, encounter date, and provider documentation are injected directly into the language model prompt as an authoritative context block before inference.

This deterministic injection strategy ensures that responses are generated using the latest available patient information rather than relying solely on retrieved semantic context, substantially reducing temporal inconsistencies and improving factual grounding.

### D. Hybrid Clinical Decision Pipeline

Following Ground Truth State Injection, semantic retrieval is performed only when additional contextual reasoning is required. Retrieved narrative documentation is combined with deterministic patient measurements before being forwarded to the selected language model.

The generated response is subsequently processed by the ProtocolManager, which maps recognized clinical terminology onto predefined evidence-based treatment pathways. Finally, the SafetyVerifier evaluates generated responses against deterministic safety rules before returning recommendations to the clinician.

This layered architecture separates deterministic computation, semantic retrieval, clinical reasoning, and safety enforcement into independent processing stages, improving transparency while reducing cascading failure modes common in monolithic LLM-based systems.

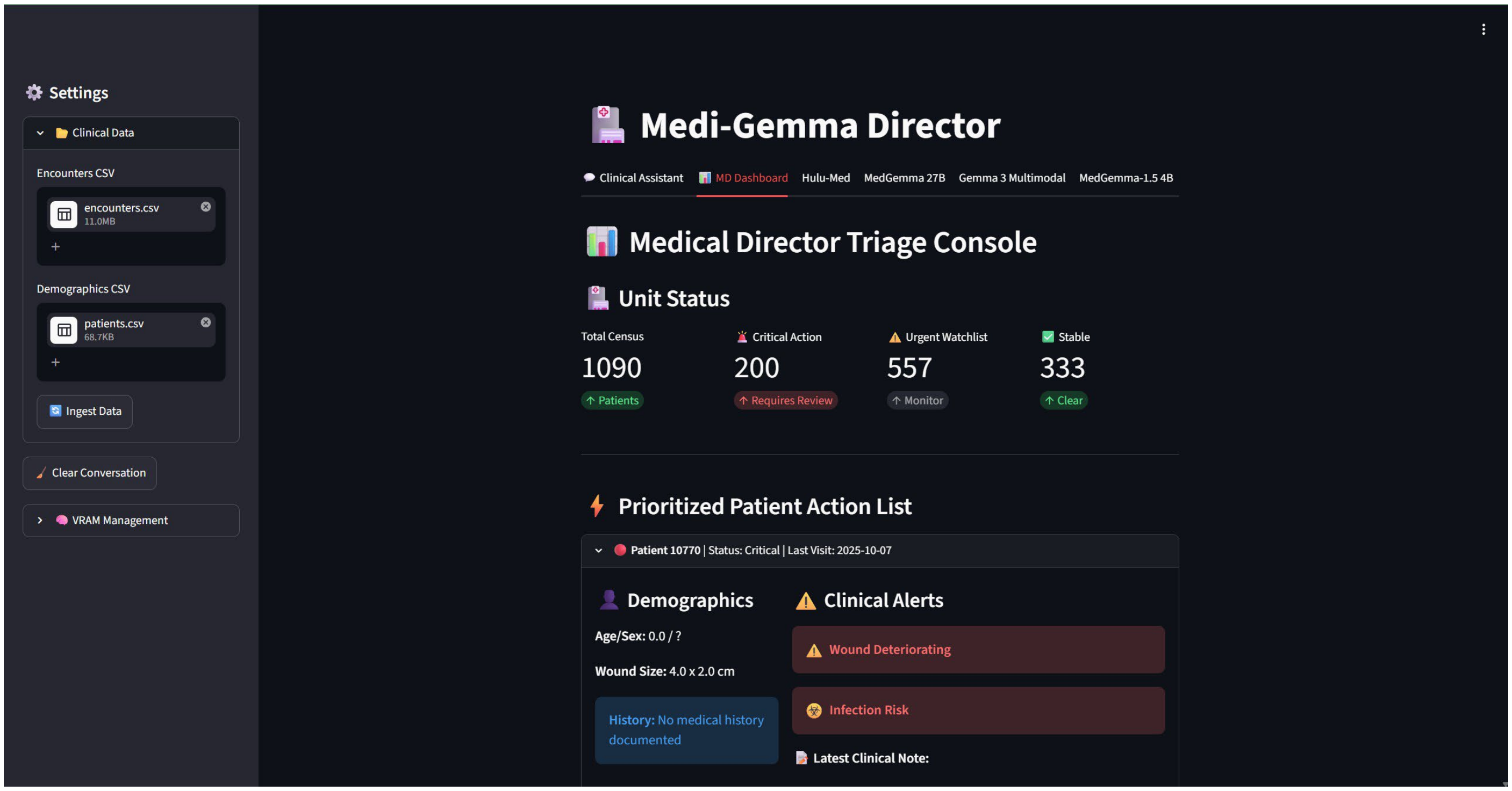

Fig. 2. Medical Director dashboard after ingestion of encounter and demographic datasets. The dashboard presents automated patient triage, wound prioritization, and longitudinal patient summaries generated by the deterministic analytics pipeline.

### E. *Representative Clinical Workflow*

Figures 1 and 3 illustrates the execution pathway followed during a typical clinician interaction. The complete workflow consists of the following stages:

1) Upload structured patient and encounter datasets.
2) Submit a natural-language clinical query.
3) IntentRouter determines whether the request requires deterministic analytics or patient-specific reasoning.
4) Statistical queries are executed directly by the PandasQueryEngine.
5) Patient-specific queries invoke the ClinicalOrchestrator for Ground Truth State Injection.
6) ClinicalRAGEngine retrieves supplementary narrative context when required.
7) ProtocolManager applies evidence-based protocol mappings.
8) SafetyVerifier validates the generated response before presentation to the clinician.

The modular design allows each subsystem to perform a specialized task while maintaining strict separation between deterministic computation and generative reasoning.

### F. *System Discussion*

The proposed architecture demonstrates that deterministic computation and retrieval-augmented generation complement one another within clinical decision support. Rather than requiring a language model to perform numerical reasoning directly over structured EMR tables, quantitative analytics are delegated to deterministic database operations while language models focus exclusively on contextual interpretation and natural language generation.

Unlike conventional retrieval pipelines that depend exclusively on vector similarity, the proposed architecture introduces a deterministic execution path for structured clinical data. Numerical calculations, patient-level analytics, temporal comparisons, and cohort statistics are executed directly through Pandas over validated EMR records, while semantic retrieval is reserved for contextual reasoning over unstructured clinical narratives. This separation significantly reduces hallucination risk during quantitative reasoning and improves reproducibility of patient-specific responses.

Ground Truth State Injection further improves factual consistency by explicitly prioritizing the latest validated patient state over semantically retrieved historical context. Together with protocol mapping and safety verification, these architectural decisions provide a transparent execution pipeline that is more suitable for safety-critical healthcare environments than conventional end-to-end retrieval-augmented generation systems [3], [15].

### G. *Computational Characteristics*

The proposed system employs a hybrid computational pipeline consisting of deterministic analytics, vector retrieval, and large language model inference. Structured EMR queries are executed using Pandas operations whose complexity scales linearly with the filtered patient records, while semantic retrieval is performed through vector similarity search over the indexed encounter database. Because deterministic analytics and retrieval operate independently, numerical reasoning remains reproducible regardless of language model behavior. This modular execution also enables individual system components to be replaced or upgraded without affecting the remaining pipeline.

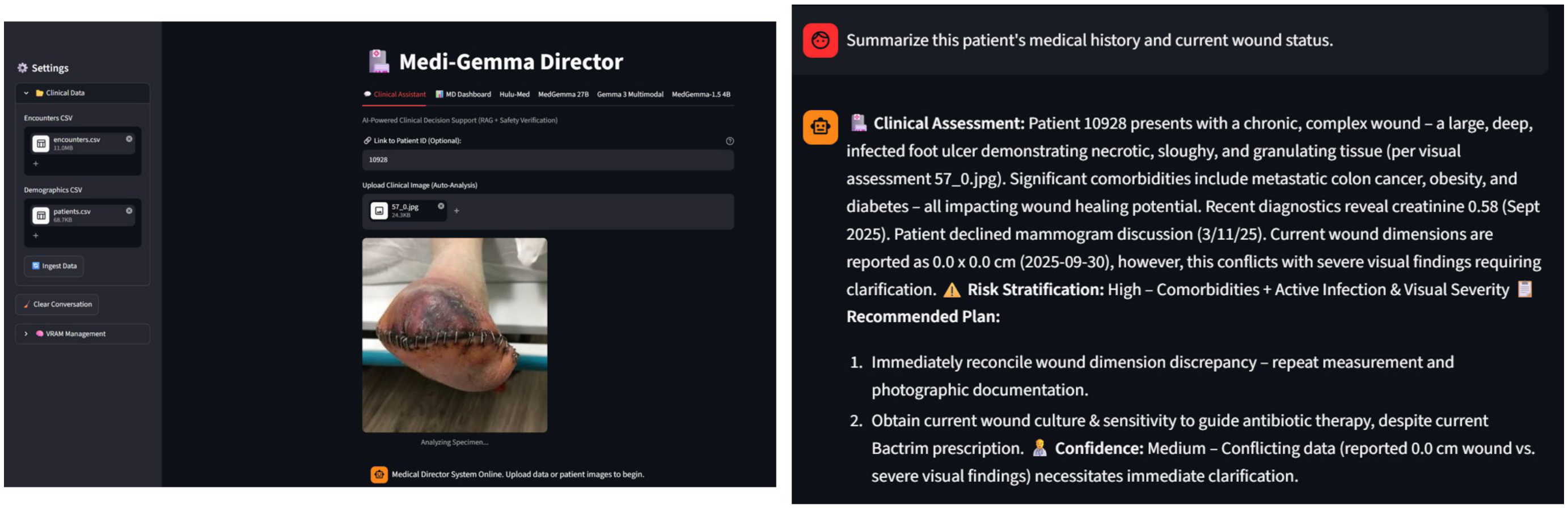

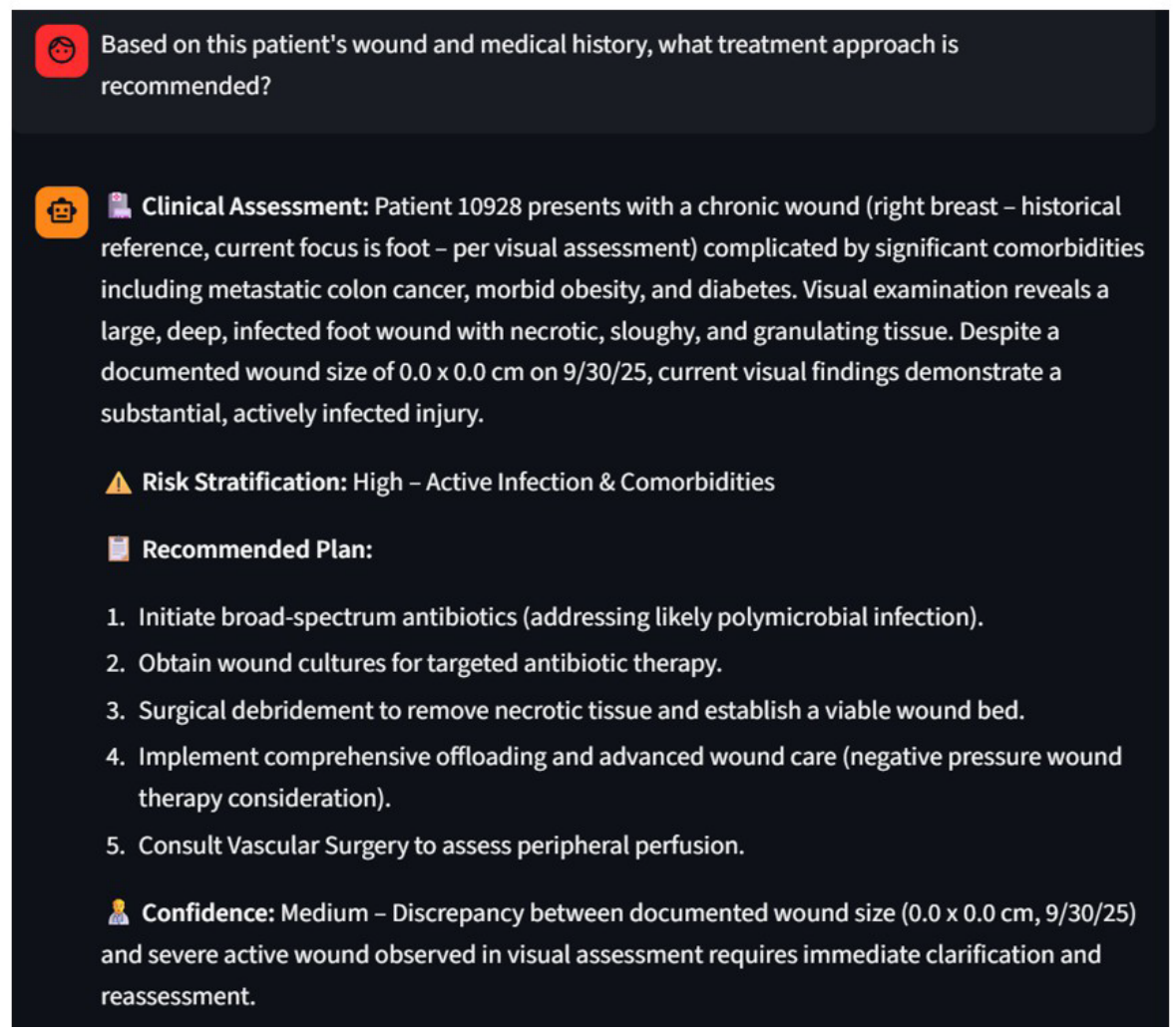

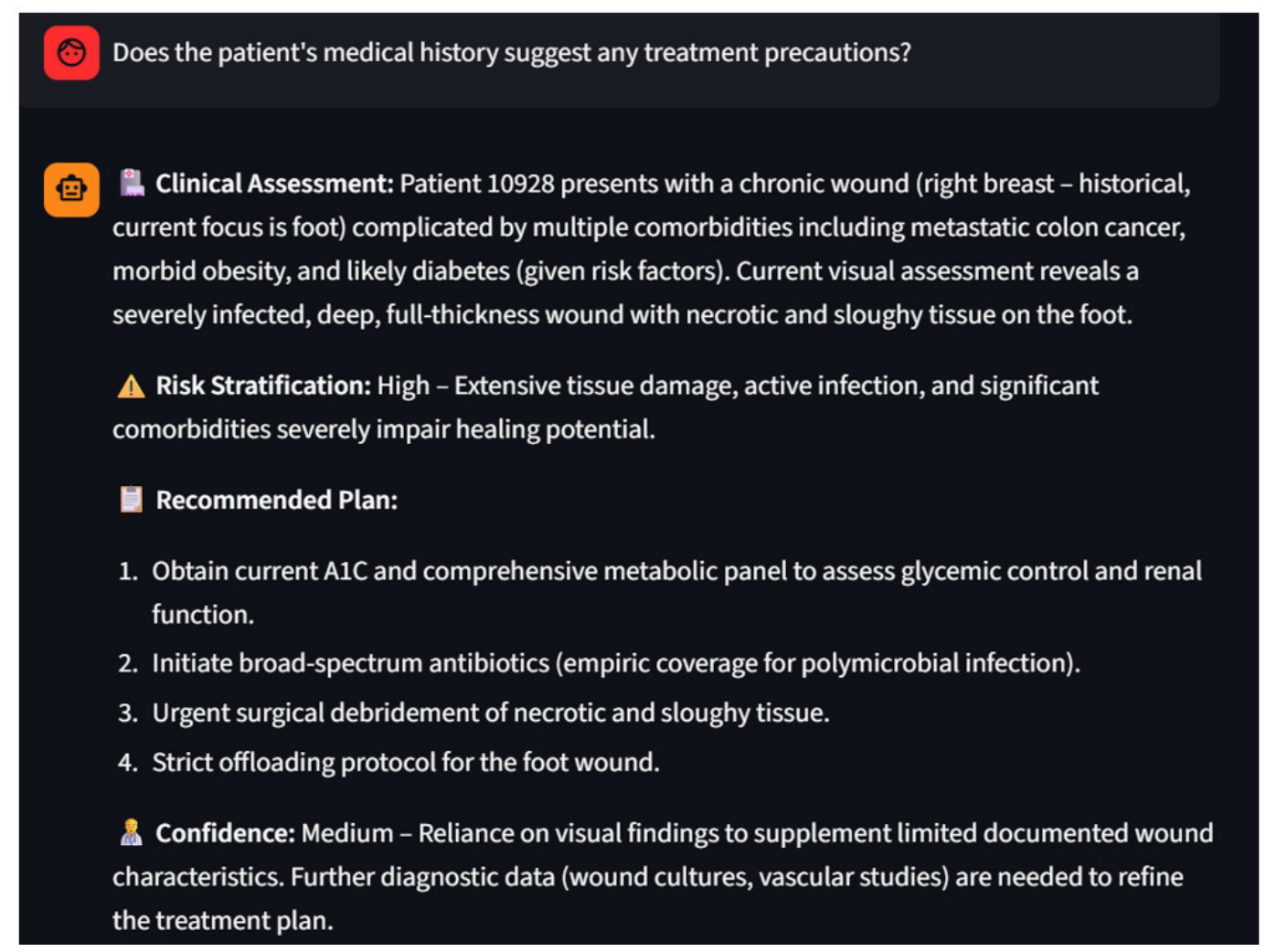

Fig. 3. Representative multimodal clinical workflow of the proposed Medi-Gemma Clinical Decision Support System. The clinician links a patient record, uploads a wound image for automatic visual analysis, and interacts with the system through natural language queries. The Clinical Orchestrator integrates deterministic patient-state retrieval, multimodal wound understanding, retrieval-augmented generation (RAG), and evidence-based protocol guidance to produce clinically grounded recommendations while maintaining factual consistency with the electronic medical record.

## V. Limitations

The proposed Clinical Decision Support System was developed as a research prototype to investigate the integration of deterministic EMR analytics with retrieval augmented language models. Although the architecture successfully demonstrates the interaction between structured data processing, semantic retrieval, and generative reasoning, several limitations remain.

First, the system has been validated using representative de-identified Electronic Medical Record (EMR) datasets and simulated clinical workflows rather than prospective deployment within a live hospital environment. Consequently, real-world usability, workflow efficiency, and clinician adoption have not yet been formally evaluated.

Second, the current implementation focuses primarily on structured wound-care documentation and associated clinical narratives. Although the architecture is generalizable to other clinical domains, additional protocol libraries, domain-specific ontologies, and specialty-specific validation would be required before deployment in broader healthcare settings. Future integration with standardized interoperability frameworks such as HL7 FHIR will facilitate deployment across heterogeneous hospital systems [11].

Third, while Ground Truth State Injection substantially improves factual grounding by incorporating the latest structured patient information, the quality of generated responses remains dependent on the completeness and accuracy of the underlying EMR data. Missing or outdated documentation may still limit downstream reasoning despite deterministic retrieval.

Finally, the current architecture employs rule-based intent routing and protocol mapping to maximize transparency and interpretability. Future versions may incorporate learned routing strategies while preserving deterministic execution for safety-critical analytical operations.

## VI. Future Work

Future development of the proposed Clinical Decision Support System will focus on expanding both its clinical capabilities and deployment readiness. First, integration with standardized Electronic Health Record (EHR) interoperability frameworks such as HL7 FHIR [11] will enable seamless communication with existing hospital information systems. Second, future versions will incorporate multimodal reasoning by combining structured EMR data with wound images, laboratory findings, and vascular imaging to provide richer clinical context.

Additional work will investigate adaptive intent routing using lightweight learned classifiers while preserving deterministic execution for safety-critical analytical tasks. We also plan to extend the protocol management framework with guideline repositories covering additional clinical specialties beyond wound care. Finally, prospective clinical evaluation with practicing healthcare professionals will assess workflow efficiency, clinician trust, diagnostic agreement, and user acceptance under real-world clinical conditions. Future work will also investigate multi-agent clinical reasoning in which specialized agents collaborate to perform longitudinal patient monitoring, protocol retrieval, risk assessment, and treatment planning while maintaining deterministic execution for structured clinical computations.

## VII. Conclusion

This paper presented Medi-Gemma, a hybrid Clinical Decision Support System that combines deterministic EMR analytics, retrieval-augmented generation, dynamic intent routing, Ground Truth State Injection, and rule-based clinical safety mechanisms within a unified agentic architecture. By explicitly separating deterministic computation from language generation, the proposed framework enables reliable processing of structured Electronic Medical Records while preserving the flexibility of natural language interaction for patient-specific reasoning.

Unlike conventional RAG pipelines that rely exclusively on semantic retrieval, the proposed architecture deterministically injects the latest validated patient state directly into the language model prompt, reducing temporal inconsistencies and improving factual grounding. Together with protocol mapping and safety verification, this modular design provides a transparent and extensible foundation for trustworthy AI-assisted clinical decision support.

Future clinical deployment and prospective validation will evaluate the effectiveness of the proposed architecture across diverse healthcare environments and clinical specialties. The modular design further provides a foundation for future multimodal clinical agents capable of jointly reasoning over structured EMRs, wound images, laboratory results, and clinical guidelines while maintaining deterministic safety guarantees required for trustworthy AI-assisted healthcare.

## System Availability

A live demonstration of the proposed Medi-Gemma Clinical Decision Support System is available through a secure web deployment:

https://unhailed-unsuccinctly-manda.ngrok-free.dev

The demonstration supports patient-data ingestion, multimodal wound-image analysis, dashboard-based clinical analytics, and natural-language clinical question answering using the complete agentic pipeline described in this work.

## References

[1] H. Touvron *et al.*, "LLaMA: Open and Efficient Foundation Language Models," *arXiv preprint arXiv:2302.13971*, 2023.

[2] A. Grattafiori *et al.*, "The Llama 3 Herd of Models," *arXiv preprint arXiv:2407.21783*, 2024.

[3] P. Lewis *et al.*, "Retrieval Augmented Generation for Knowledge-Intensive NLP Tasks," *Advances in Neural Information Processing Systems (NeurIPS)*, vol. 33, 2020.

[4] K. Singhal *et al.*, "Large Language Models Encode Clinical Knowledge," *Nature*, vol. 620, no. 7972, pp. 172–180, 2023.

[5] K. Singhal *et al.*, "Towards Expert-Level Medical Question Answering with Large Language Models," *arXiv preprint arXiv:2305.09617*, 2023.

[6] C. Li *et al.*, "LLaVA-Med: Training a Large Language-and-Vision Assistant for Biomedicine in One Day," *arXiv preprint arXiv:2306.00890*, 2023.

[7] Y. Kabak *et al.*, "FHIR-RAG-MEDS: Integrating HL7 FHIR with retrieval augmented Large Language Models for Enhanced Medical Decision Support," *arXiv preprint arXiv:2509.07706*, 2025.

[8] J. C. L. Ong *et al.*, "Development and Testing of a Novel Large Language Model-Based Clinical Decision Support System for Medication Safety," *arXiv preprint arXiv:2402.01741*, 2024.

[9] C. F. Kurz *et al.*, "Benchmarking Vision-Language Models for Diagnostics in Emergency and Critical Care Settings," *npj Digital Medicine*, vol. 8, no. 1, 2025.

[10] Health Level Seven International, "FHIR Release 5 Specification," 2023.

[11] J. Johnson, M. Douze, and H. Je´gou, "Billion-scale Similarity Search with GPUs," *arXiv preprint arXiv:1702.08734*, 2017.

[12] LlamaIndex, "LlamaIndex Documentation." Available: https://docs.llamaindex.ai/ Accessed: 2026.

[13] LangChain, "LangChain Documentation." Available: https://docs.langchain.com/ Accessed: 2026.

[14] Y. Wang *et al.*, "A Survey on Large Language Model Based Autonomous Agents," *arXiv preprint arXiv:2308.11432*, 2023.

[15] W. McKinney, "Data Structures for Statistical Computing in Python," *Proceedings of the 9th Python in Science Conference*, pp. 51–56, 2010.

[16] Y. Gao, Y. Xiong, X. Gao, et al., "Retrieval-Augmented Generation for Large Language Models: A Survey," *arXiv preprint arXiv:2312.10997*, 2024.

[17] M. Moor, O. Banerjee, Z. S. H. Abad, H. M. Krumholz, J. Leskovec, E. J. Topol, and P. Rajpurkar, "Foundation Models for Generalist Medical Artificial Intelligence," *Nature*, vol. 616, no. 7956, pp. 259–265, Apr. 2023, doi: 10.1038/s41586-023-05881-4.

[18] S. Yao *et al.*, "ReAct: Synergizing Reasoning and Acting in Language Models," *International Conference on Learning Representations (ICLR)*, 2023.

[19] T. Schick *et al.*, "Toolformer: Language Models Can Teach Themselves to Use Tools," *Advances in Neural Information Processing Systems (NeurIPS)*, 2023.

[20] Y. Shen *et al.*, "HuggingGPT: Solving AI Tasks with ChatGPT and its Friends in Hugging Face," *Advances in Neural Information Processing Systems (NeurIPS)*, 2023.

[21] World Health Organization, "Ethics and Governance of Artificial Intelligence for Health," WHO, Geneva, 2021.

[22] T. Singhal *et al.*, "Towards Safe and Reliable Large Language Models for Medicine," *arXiv preprint arXiv:2403.03744v2*, 2024.